\documentclass[11pt,a4paper]{article}
\usepackage[hyperref]{emnlp2020}
\usepackage{times}
\usepackage{latexsym}

\usepackage{graphicx}
\usepackage{microtype}

\aclfinalcopy 
\def\aclpaperid{33} 

\title{The elephant in the interpretability room:\\
Why use attention as explanation when we have saliency methods?}

\author{Jasmijn Bastings\\
Google Research\\
  \texttt{bastings@google.com} \\\And
  Katja Filippova \\
    Google Research\\
  \texttt{katjaf@google.com}}

\date{}

\begin{document}
\maketitle
\begin{abstract}
There is a recent surge of interest in using attention as explanation of model predictions, with mixed evidence on whether attention can be used as such. While attention conveniently gives us one weight per input token and is easily extracted, it is often unclear toward what goal it is used as explanation. We find that often that goal, whether explicitly stated or not, is to find out what input tokens are the most relevant to a prediction, and that the implied user for the explanation is a model developer. For this goal and user, we argue that input saliency methods are better suited, and that there are no compelling reasons to use attention, despite the coincidence that it provides a weight for each input. With this position paper, we hope to shift some of the recent focus on attention to saliency methods, and for authors to clearly state the goal and user for their explanations.
\end{abstract}

\section{Introduction}
\label{sec:introduction}
Attention mechanisms \citep{bahdanau-et-al-2015-neural} have allowed for performance gains in many areas of NLP, including, \emph{inter alia}, machine translation \citep{bahdanau-et-al-2015-neural,luong-etal-2015-effective,vaswani-et-al-2017-attention}, 
natural language generation \citep[e.g.,][]{rush15,narayan-xsum}, and natural language inference \citep[e.g.,][]{parikh-etal-2016-decomposable}. 

Attention has not only allowed for better performance, it also provides a window into how a model is operating. For example, for machine translation, \citet{bahdanau-et-al-2015-neural}  visualize what source tokens the target tokens are attending to, often aligning words that are translations of each other.

Whether the window that attention gives into how a model operates amounts to \emph{explanation} has recently become subject to debate (\S\ref{sec:background}). While many papers published on the topic of explainable AI have been criticised for not defining explanations \citep{lipton-mythos-2018,miller-xai-insights-2019}, the first key studies which spawned interest in attention as explanation  \citep[][]{jain-wallace-2019-attention,serrano-smith-2019-attention,wiegreffe-pinter-2019-attention} do say that they are interested in whether attention weights faithfully represent the responsibility each input token has on a model prediction. 
That is, the narrow definition of explanation implied there is that it points at the most important input tokens for a prediction (arg max), accurately summarizing the reasoning process of the model \citep{jacovi-goldberg-2020-towards}.

The above works have inspired some to find ways to make attention  more faithful and/or plausible, by changing the nature of the hidden representations attention is computed over using special training objectives \citep[e.g.,][]{mohankumar-etal-2020-towards, tutek-snajder-2020-staying}. Others have proposed replacing the attention mechanism with a latent alignment model \citep{deng-etal-2018-latent}.

Interestingly, the implied definition of explanation in the cited works, happens to coincide with what \textit{input saliency methods} (\S\ref{sec:saliency-methods}) are designed to produce \citep[][i.a.]{li-etal-2016-visualizing,sundararajan-ig-2017, ribeiro-lime, montavon-lrp-overview-2019}. 
Moreover, the user of that explanation is often implied to be a model developer, to which faithfulness is important.
The elephant in the room is therefore:
If the goal of using attention as explanation is to assign importance weights to the input tokens in a faithful manner, why should the attention mechanism be preferred over the multitude of existing input saliency methods designed to do \emph{exactly that}? 
In this position paper, with that goal in mind, we argue that we should pay attention no heed (\S\ref{sec:core-argument}). 
We propose that we reduce our focus on attention as explanation, and shift it to input saliency methods instead. 
However, we do emphasize that understanding the \emph{role} of attention is still a valid research goal (\S\ref{sec:not-not-interesting}), and finally, we discuss a few approaches that go beyond saliency (\S\ref{sec:beyond-saliency}).

\clearpage

\section{The Attention Debate}
\label{sec:background}

\begin{figure}[t]
    \centering
    \includegraphics[width=5cm,clip,trim=0 0 0 2mm]{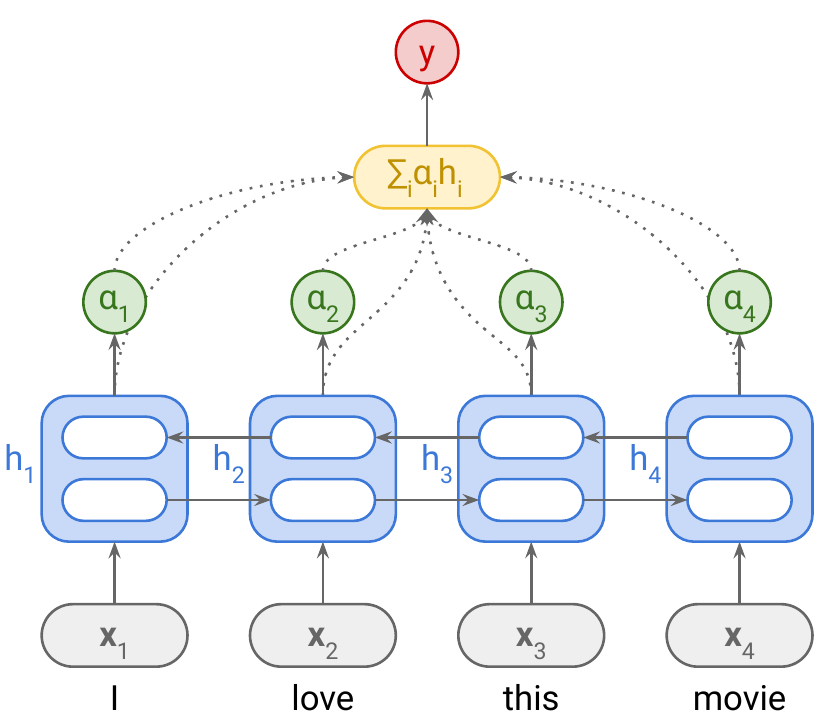}
    \caption{A typical model in the debate.}
    \label{fig:model}
\end{figure}

In this section we summarize the debate on whether attention is explanation. The debate mostly features simple BiLSTM text classifiers (see Figure~\ref{fig:model}). Unlike Transformers \citep{vaswani-et-al-2017-attention}, they only contain a single attention mechanism, which is typically MLP-based \citep{bahdanau-et-al-2015-neural}:
\begin{equation}
  \!  e_i \!=\! \mathbf{v}^\top \mathrm{tanh}\big(W_h \mathbf{h}_i \!+\! W_q \mathbf{q}\big) \label{eq:att} \,\,\, \alpha_i = \frac{\exp e_i}{\sum_k \exp e_k}
\end{equation}
\noindent where $\alpha_i$ is the attention score for BiLSTM state $\mathbf{h}_i$. 
When there is a single input text, there is no query, and $\mathbf{q}$ is either a trained parameter (like $\mathbf{v}, W_h$ and $W_q$), or $W_q \mathbf{q}$ is simply left out of Eq.~\ref{eq:att}.

\subsection{Is attention (not) explanation?}
\citet{jain-wallace-2019-attention} show that attention is often uncorrelated with gradient-based feature importance measures, and that one can often find a completely different set of attention weights that results in the same prediction. In addition to that, \citet{serrano-smith-2019-attention} find, by modifying attention weights, that they often do not identify those representations that are most most important to the prediction of the model. 
However, \citet{wiegreffe-pinter-2019-attention} claim that these works do not disprove the usefulness of attention as explanation \emph{per se}, and provide four tests to determine if or when it can be used as such. In one such test, they are able to find alternative attention weights using an adversarial training setup, which suggests attention is not always a faithful explanation.
Finally, \citet{pruthi-etal-2020-learning} propose a method to produce deceptive attention weights. Their method reduces how much weight is assigned to a set of `impermissible' tokens, even when the models demonstratively rely on those tokens for their predictions.

\label{sec:problems}
\subsection{Was the right task analyzed?}
In the attention-as-explanation research to date text classification with LSTMs received the most scrutiny. However, \citet{vashishth-et-al-2019-attention-interpretability} question why one should focus on single-sequence tasks at all because the attention mechanism is arguably far less important there than in models involving two sequences, like NLI or MT. Indeed, the performance of an NMT model degrades substantially if uniform weights are used, while random attention weights affect the text classification performance minimally. Therefore, findings from text classification studies may not generalize to tasks where attention is a crucial component. 

Interestingly, even for the task of MT, the first case where attention was visualized to inspect a model (\S\ref{sec:introduction}), \citet{ding-etal-2019-saliency} find that saliency methods (\S\ref{sec:saliency-methods}) yield better word alignments.

\subsection{Is a causal definition assumed?}
\citet{grimsley-etal-2020-attention} go as far as saying that attention is not explanation by definition, if a causal definition of explanation is assumed. Drawing on the work in philosophy, they point out that causal explanations presuppose that a surgical intervention is possible which is not the case with deep neural networks: one cannot intervene on attention while keeping all the other variables invariant. 

\subsection{Can attention be improved?}
The problems with using as attention as explanation, especially regarding faithfulness, have inspired some to try and `improve' the attention weights, so to make them more faithful and/or plausible.
\citet{mohankumar-etal-2020-towards} observe high similarity between the hidden representations of LSTM states and propose a diversity-driven training objective that makes the hidden representations more diverse across time steps. They show using representation erasure that the resulting attention weights result in decision flips more easily as compared to vanilla attention.
With a similar motivation, \citet{tutek-snajder-2020-staying} use a word-level objective to achieve a stronger connection between hidden states and the words they represent, which affects attention.
Not part of the recent debate, \citet{deng-etal-2018-latent} propose variational attention as an alternative to the soft attention of \citet{bahdanau-et-al-2015-neural}, arguing that the latter is not \emph{alignment}, only an approximation thereof. They have the additional benefit of allowing posterior alignments, conditioned on the input and the output sentences.

\section{Saliency Methods}
\label{sec:saliency-methods}
In this section  we discuss various input saliency methods for NLP as alternatives to attention: gradient-based (\S\ref{sec:gradient-based}), propagation-based (\S\ref{sec:lrp}), and occlusion-based methods (\S\ref{sec:occlusion-based}),  following \citet{arras-etal-2019-evaluating}.
We do not endorse any specific method\footnote{For an evaluation of methods for explaining LSTM-based models, see e.g.,  \citet{poerner-etal-2018-evaluating} and \citet{arras-etal-2019-evaluating}. 
}, 
but rather try to give an overview of methods and how they differ.
We discuss methods that are applicable to \emph{any} neural NLP model, allowing access to model internals, such as activations and gradients, as attention itself requires such access. We leave out more expensive methods that use a surrogate model, e.g., LIME \citep{ribeiro-lime}.

\subsection{Gradient-based methods}
\label{sec:gradient-based}
While used earlier in other fields,
\citet{li-etal-2016-visualizing} use gradients as explanation in NLP and compute:
\begin{equation}
    \nabla_{\mathbf{x}_i} f_c(\mathbf{x}_{1:n})\label{eq:grad}
\end{equation}
where $\mathbf{x}_i$ is the input word embedding for time step $i$, $\mathbf{x}_{1:n} = \langle \mathbf{x}_1, \dots, \mathbf{x}_n\rangle$ are the input embeddings (e.g., a sentence), and $f_c(\mathbf{x}_{1:n})$ the model output for target class $c$. After taking the L2 norm of Eq.~\ref{eq:grad}, the result is a measure of how sensitive the model is to the input at time step $i$.

If instead we take the dot product of Eq.~\ref{eq:grad} with the input word embedding $\mathbf{x}_i$, we arrive at the gradient$\times$input method \citep{denil-et-al-2015-extraction}, which returns a saliency (scalar) of input $i$:
\begin{equation}
    \nabla_{\mathbf{x}_i} f_c(\mathbf{x}_{1:n}) \cdot \mathbf{x}_i
\end{equation}

Integrated gradients (IG) \citep{sundararajan-ig-2017} is a gradient-based method which deals with the problem of \emph{saturation}: gradients may get close to zero for a well-fitted function. IG requires a baseline $\mathbf{b}_{1:n}$, e.g., all-zeros vectors or repeated [MASK] vectors. 
For input $i$, we compute:
\begin{equation}
    \frac{1}{m} \!\! \sum_{k=1}^m \!\!\nabla_{\mathbf{x}_i} f_c\!\Big(\!\mathbf{b}_{1:n}\!\!+\! \frac{k}{m} \! (\mathbf{x}_{1:n}\!-\!\mathbf{b}_{1:n} )\!\Big) \!\cdot\!  (\mathbf{x}_i \!-\! \mathbf{b}_i) \!
\end{equation}
\noindent That is, we average over $m$ gradients, with the inputs to $f_c$ being linearly interpolated between the baseline and the original input $\mathbf{x}_{1:n}$ in $m$ steps. We then take the dot product of that averaged gradient with the input embedding $\mathbf{x}_i$ minus the baseline.

We propose distinguishing \emph{sensitivity} from \emph{saliency}, following \citet{ancona-et-al-2019-gradient}: the former says how much a change in the input changes the output, while the latter is the marginal effect of each input word on the prediction. Gradients measure sensitivity, whereas gradient$\times$input and IG measure saliency.
A model can be sensitive to the input at a time step, but it depends on the actual input vector if it was important for the prediction.

\subsection{Propagation-based methods}
\label{sec:lrp}
Propagation-based methods \citep[][i.a.]{landecker-etal-2012-interpreting,bach-et-al-2015-lrp,arras-etal-2017-explaining}, of which we discuss Layer-wise Relevance Propagation (LRP) in particular, start with a forward pass to obtain the output $f_c(\mathbf{x}_{1:n})$, which is the top-level \emph{relevance}. They then use a special backward pass that, at each layer, \emph{redistributes} the incoming relevance among the inputs of that layer.
Each kind of layer has its own propagation rules. For example, there are different rules for feed-forward layers \citep[][]{bach-et-al-2015-lrp} and LSTM layers \citep{arras-etal-2017-explaining}.
Relevance is redistributed until we arrive at the input layers.
While LRP requires implementing a custom backward pass, 
it does allow precise control to preserve relevance,
and it has been shown to work better than using gradient-based methods on text classification \citep{arras-etal-2019-evaluating}.

\subsection{Occlusion-based methods}
\label{sec:occlusion-based}
Occlusion-based methods \citep{zeiler-fergus-2014-visualizing,li-etal-2016-understanding} compute input saliency by occluding (or erasing) input features and measuring how that affects the model.
Intuitively, erasing unimportant features does not affect the model, whereas the opposite is true for important features.
\citet{li-etal-2016-understanding} erase word embedding dimensions and whole words to see how doing so affects the model. 
They compute the importance of a word \emph{on a dataset} by averaging over how much, for each example, erasing that word caused a difference in the output compared to not erasing that word.

As a saliency method, however, we can apply their method on a single example only. For input $i$:
\begin{equation}
    f_c(\mathbf{x}_{1:n}) - f_c(\mathbf{x}_{1:n \mid \mathbf{x}_i = 0})
\end{equation}
\noindent computes saliency, where $\mathbf{x}_{1:n \mid \mathbf{x}_i = 0}$ indicates that input word embedding $\mathbf{x}_i$ was zeroed out, while the other inputs were unmodified.
\citet{kadar-etal-2017-representation} and \citet{poerner-etal-2018-evaluating} use a variant, \emph{omission}, by simply leaving the word out of the input.

This method requires $n+1$ forward passes. It is also used for evaluation, to see if important words another method has identified bring a change in model output \citep[e.g.,][]{deyoung-etal-2020-eraser}. 

\section{Saliency vs. Attention}
\label{sec:core-argument}
We discussed the use of attention as explanation (\S\ref{sec:background}) and input saliency methods as alternatives (\S\ref{sec:saliency-methods}). We will now argue why saliency methods should be preferred over attention for explanation.

In many of the cited papers, whether implicitly or explicitly, the \emph{goal} of the explanation is to reveal which input words are the most important ones for the final prediction. 
This is perhaps a consequence of attention computing one weight per input, so it is necessarily understood in terms of those inputs.

The intended \emph{user} for the explanation is often not stated, but typically that user is a model developer, and not a non-expert end user, for example. 
For model developers, faithfulness, the need for an explanation to accurately represent the reasoning of the model, is a key concern. On the other hand, plausibility is of lesser concern, because a model developer aims to understand and possibly improve the model, and that model does not necessarily align with human intuition \citep[see][for a detailed discussion of the differences between faithfulness and plausibility]{jacovi-goldberg-2020-towards}.  

With this goal and user clearly stated, it is impossible to make an argument in favor of using attention as explanation. Input saliency methods are addressing the goal head-on: they reveal why one particular model prediction was made in terms of how relevant each input word was to that prediction. Moreover, input saliency methods typically take the entire computation path into account, all the way from the input word embeddings to the target output prediction value. Attention weights do not: they reflect, at one point in the computation, how much the model attends to each input \emph{representation}, but those representations might already have mixed in information from other inputs.
Ironically, attention-as-explanation is sometimes evaluated by comparing it against gradient-based measures, which again begs the question why we wouldn't use those measures in the first place.

One might argue that attention, despite its flaws, is easily extracted and computationally efficient. However, it only takes one line in a framework like TensorFlow to compute the gradient of the output w.r.t. the input word embeddings, so implementation difficulty is not a strong argument. In terms of efficiency, it is true that for attention only a forward pass is required, but many other methods discussed at most require a forward and then a backward pass, which is still extremely efficient.

\section{Attention is not not interesting}
\label{sec:not-not-interesting}
In this position paper we criticized the use of attention to assess input saliency for the benefit of the model developer.
We emphasize that understanding the \textit{role} of the attention mechanism is a perfectly justified research goal. 
For example, \citet{voita-etal-2019-analyzing} and \citet{michel-attention-heads-2019} analyze the role of attention heads in the Transformer architecture and identify a few distinct functions they have, and \citet{strubell-etal-2018-linguistically} train attention heads to perform dependency parsing, adding a linguistic bias.

We also stress that if the definition of explanation is adjusted, for example if a different intended \textit{user} and a different explanatory \textit{goal} are articulated, attention may become a useful explanation for a certain application. 
For example, \citet{strout-etal-2019-human} demonstrate that supervised attention helps humans accomplish a task faster than random or unsupervised attention, for a user and goal that are very different from those implied in \S\ref{sec:background}. 

\section{Is Saliency the Ultimate Answer?}
\label{sec:beyond-saliency}
\paragraph{Beyond saliency.}
While we have argued that saliency methods are a good fit for our goal, there are other goals for which different methods can be a better fit.
For example, counterfactual analysis might lead to insights, aided by visualization tools  \citep{vig-2019-multiscale,hoover-etal-2020-exbert,abnar-zuidema-2020-quantifying}.
The DiffMask method of \citet{decao-diffmask-2020}  adds another dimension: it not only reveals in what layer a model knows what inputs are important, but also where important information is stored as it flows through the layers of the model.
Other examples are models that rationalize their predictions \citep{lei-etal-2016-rationalizing,bastings-etal-2019-interpretable}, which can guarantee faithful explanations, although they might be sensitive to so-called \emph{trojans} \citep{jacovi2020aligning}.

\paragraph{Limitations of saliency.}
A known problem with occlusion-based saliency methods as well as erasure-based evaluation of any input saliency technique \cite{bach-et-al-2015-lrp,deyoung-etal-2020-eraser} is that changes in the predicted probabilities may be due to the fact that the corrupted input falls off the manifold of the training data \cite{hooker-benchmark}. That is, a drop in probability can be explained by the input being OOD and not by an important feature missing. It has also been demonstrated that at least some of the saliency methods are not reliable and produce unintuitive results \cite{kindermans-unreliability} or violate certain axioms \cite{sundararajan-ig-2017}. 

A more fundamental limitation is the expressiveness of input saliency methods. Obviously, a bag of per-token saliency weights can be called an explanation only in a very narrow sense. One can overcome some limitations of the flat representation of importance by indicating dependencies between important features (for example, \citet{janizek-explaining-explanations} present an extension of IG which explains pairwise feature interactions) but it is hardly possible to fully understand why a deep non-linear model produced a certain prediction by only looking at the input tokens. 

\section{Conclusion}
\label{sec:conclusion}
We summarized the debate on whether attention is explanation, and observed that the goal for explanation is often to determine what inputs are the most relevant to the prediction. The user for that explanation often goes unstated, but is typically assumed to be a model developer. With this goal and user clearly stated, we argued that input saliency methods---of which we discussed a few---are better suited than attention. We hope, at least for the goal and user that we identified, that the focus shifts from attention to input saliency methods, and perhaps to entirely different methods, goals, and users. 

\section*{Acknowledgments}

We would like to thank Sebastian Gehrmann for useful comments and suggestions, as well as our anonymous reviewers, one of whom mentioned there is a whale in the room as well.

\bibliography{main}
\bibliographystyle{acl_natbib}

\end{document}